\documentclass{article} 
\usepackage{iclr2027_conference,times}

\usepackage{amsmath,amsfonts,bm}

\def\eqref#1{equation~\ref{#1}}

\def\1{\bm{1}}

\DeclareMathAlphabet{\mathsfit}{\encodingdefault}{\sfdefault}{m}{sl}
\SetMathAlphabet{\mathsfit}{bold}{\encodingdefault}{\sfdefault}{bx}{n}

\usepackage{hyperref}
\usepackage{url}
\usepackage{graphicx} 
\graphicspath{{./includes/}, {./includes/supplementary_includes/}}
\usepackage[]{units} 
\usepackage{xspace} 
\usepackage{amsfonts} 
\usepackage[dvipsnames]{xcolor} 
\usepackage{subcaption}  
\usepackage{float}
\usepackage{amsmath} 
\usepackage{bbm}     
\usepackage[T1]{fontenc}
\usepackage{algorithm}
\usepackage{algpseudocode}
\usepackage{amssymb}
\usepackage{booktabs}
\usepackage{wrapfig}
\usepackage{enumitem}

\usepackage[commandnameprefix=ifneeded]{changes}
\definechangesauthor[name={Sehoon Ha}, color=blue]{SH}
\setdeletedmarkup{}              
\renewcommand{\deleted}[2][]{}   

\definechangesauthor[name={Wontaek Kim}, color=orange]{WK}

\newcommand{\methodabbr}{PrefPI}

\title{PrefPI: Preference-Guided Steering into \\ Out-of-Distribution Behaviors}

\author{Seungeun Rho\thanks{Co-first authors.} \qquad
Wontaek Kim\footnotemark[1] \qquad
Danfei Xu \qquad
Sehoon Ha \\[0.5em]
School of Interactive Computing \\
Georgia Institute of Technology \\[0.3em]
\texttt{\{srho31,wkim345,danfei,sehoonha\}@gatech.edu}
}

\iclrfinalcopy 
\begin{document}

\maketitle

\begin{abstract}
We present \textbf{\methodabbr{} (\emph{Preference-Guided Policy Iteration})}, an iterative framework for steering pretrained generative robot policies using only relative preferences over self-generated trajectories. Unlike prior preference-learning methods that primarily sharpen modes already represented by the policy, we study \emph{steering beyond the initial effective support}, where desired behaviors are rarely or never observed under the initial policy. Our key idea is to formulate preference learning as preference-conditioned generative modeling: preferred trajectories define a conditional distribution, whose density ratio with the broader behavior prior provides an implicit preference signal amplified by classifier-free guidance (CFG). Repeating this preference-conditioned modeling and guidance step yields a form of preference-guided policy iteration, turning incremental improvements toward previously inaccessible behaviors. Across diffusion policies and the PI0.5 flow-matching VLA in simulation and the real world, \methodabbr{} produces substantial behavioral shifts with limited feedback. In particular, \methodabbr{} increases object transport height from 10.7~cm to 19.8~cm  on real hardware with only 150 preference-labeled trajectories.

\end{abstract}

\begin{figure*}[h!]
    \centering
    \includegraphics[width=1.0\textwidth]{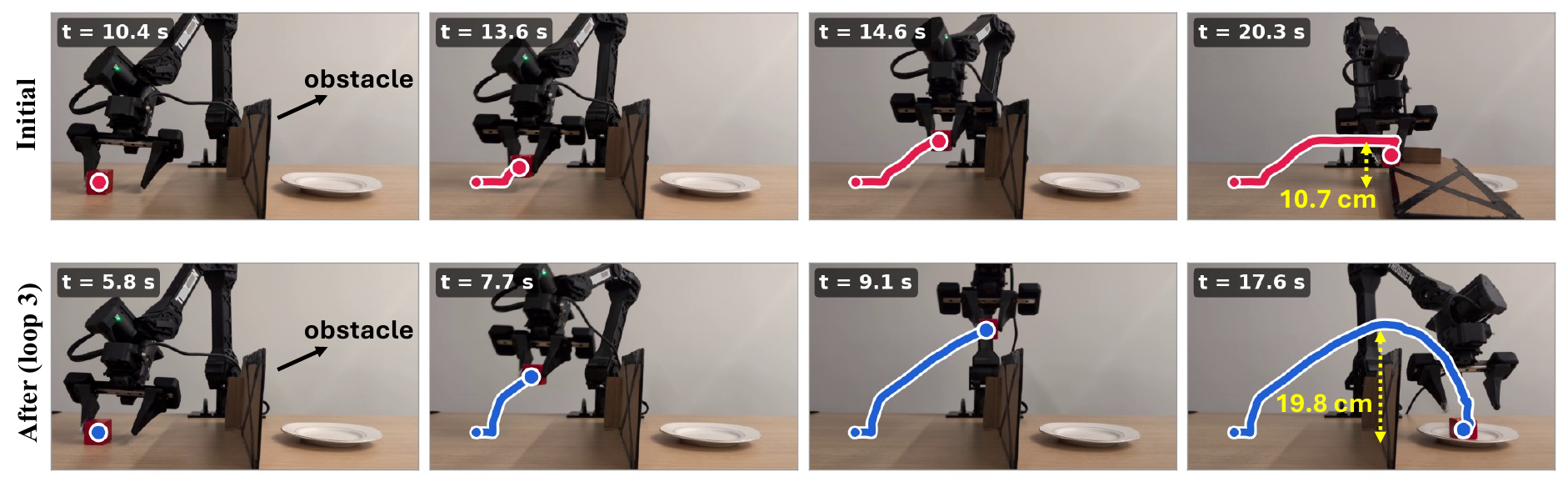}
    \vspace{-2em} 
    \caption{With only \textbf{150} preference-labeled trajectories (50 per loop over 3 loops), \methodabbr{} increased the object’s maximum height from 10.7 cm to 19.8cm on real hardware, enabling obstacle clearance.}
    \label{fig:teaser}
\end{figure*}

\section{Introduction}

Pretrained generative robot policies are increasingly capable of solving a wide range of manipulation tasks, yet their behavior may not match the preferences of individual users. A robot may already know how to complete a task, while a user prefers that it move faster, follow a different trajectory, or achieve a different final configuration. Adapting such policies after deployment therefore requires a lightweight mechanism for steering an already capable policy toward user-preferred behaviors without collecting new demonstrations or manually specifying reward functions.

Preference feedback provides a natural interface for this purpose. Users can evaluate trajectories generated by the robot itself rather than demonstrate the desired behavior. Prior work has shown that preference-based adaptation can improve task success, suppress undesirable behaviors, and align pretrained policies with user preferences~\citep{biyik2018batch,hejna2023fewshot,metcalf2023sample,chen2025fdpp,wallace2024diffusion,wu2026flowpro}. However, these evaluations largely optimize behaviors that are already represented, or at least observable, under the policy during data collection. We study a more challenging setting: can preference feedback steer a policy toward desirable behaviors that \emph{cannot be produced} by the initial policy?

We call this setting \textbf{steering beyond the initial effective support}.%
\footnote{We use \emph{effective support} to denote the set of trajectories whose density is at least a small threshold $\epsilon>0$, i.e., \mbox{$\{\tau \mid p(\tau)\geq\epsilon\}$}.} If a desired behavior lies outside this set, it is exceedingly unlikely under the initial policy, may never appear in the first rollout batch, and therefore cannot directly supervise a single-round preference update. An iterative procedure can instead use incremental improvements as stepping stones, with each update shifting the rollout distribution, exposing new behaviors that can be preferred in subsequent rounds.

To realize this idea, we introduce \textbf{\methodabbr{} (\emph{Preference-Guided Policy Iteration})}, an iterative framework for steering pretrained generative robot policies using only relative preferences over self-generated trajectories. Our key idea is to formulate preference learning as \emph{preference-conditioned generative modeling}. Rather than learning an explicit reward function or directly optimizing preference pairs, we treat preferred trajectories as samples from a conditional distribution and the broader behavior distribution as an unconditional reference. Their density ratio then provides an implicit preference signal that classifier-free guidance (CFG)~\citep{ho2022classifierfree} uses to bias generation toward preferred behaviors.

At each iteration, the current policy generates a batch of rollouts and the user selects the relatively preferred subset. We train a preference-conditioned branch on these selected trajectories and an unconditional branch on an accumulated set of successful trajectories. CFG combines the two branches to produce the next policy, biasing generation toward the current preference while retaining a broad behavioral prior. Repeating this preference-conditioned modeling and guidance step yields a form of preference-guided policy iteration, allowing progressively more preferred behaviors. We derive that top-$m$ selection induces an order-preserving preference signal, which CFG amplifies to improve expected latent utility.

We evaluate \methodabbr{} across simulated and real-world manipulation tasks with both diffusion policies and the PI0.5 flow-matching VLA~\citep{pmlr-v305-black25a}. In simulation, \methodabbr{} increases detour height from approximately 15~cm to 37~cm and horizontal displacement from 4~cm to 12~cm, producing trajectories that extend well beyond regions visited by the initial policy. On real robot hardware, object transport height increases from 10.7~cm to 19.8~cm after only three iterations, and further reaches 20.4~cm after six iterations. \methodabbr{} also reduces average task-completion time from 19.8~s to 10.8~s and increases throwing distance by $4.5\times$, while maintaining above $90\%$ task success.

Our contributions are:
\begin{itemize}
\item We introduce a challenging preference-learning setting, \emph{steering beyond the initial effective support}, where the desired behavior is extremely unlikely to appear under the initial policy distribution and no additional demonstrations are provided.

\item We propose \methodabbr{}, an iterative framework that uses preference-conditioned generative modeling and CFG to progressively steer toward such behaviors.

\item We show that relative top-$m$ selection induces an order-preserving optimality signal, enabling CFG to serve as a policy-improvement operator.

\item Across diffusion and flow-matching policies in simulation and the real world, we demonstrate that \methodabbr{} achieves large behavioral shifts with limited preference feedback.

\end{itemize}

\section{Related Work}

\subsection{Preference-Based Robot Policy Adaptation}

Preference-based robot learning reduces the need for demonstrations and hand-designed rewards by allowing users to compare candidate behaviors. Early work learns reward functions from pairwise trajectory comparisons and uses active querying to reduce human feedback~\citep{biyik2018batch}. Later methods improve sample and feedback efficiency through active data collection, relabeling, and exploratory pretraining~\citep{lee2021pebble,hejna2023fewshot,metcalf2023sample}.

Recent methods adapt pretrained generative robot policies directly from preference feedback. FDPP~\citep{chen2025fdpp} learns a preference reward from policy-generated trajectories and fine-tunes a diffusion policy with reinforcement learning. Diffusion-DPO~\citep{wallace2024diffusion} directly optimizes diffusion models from preferred and rejected samples, while FlowPRO~\citep{wu2026flowpro} develops a preference objective for flow-matching vision-language-action policies and performs repeated real-world data collection and fine-tuning.

Our work asks a complementary question. Rather than only increasing the probability of preferred behaviors observed during adaptation, we study whether iterative preference feedback can move a policy toward behavioral regimes that are exceedingly unlikely under its initial distribution. \methodabbr{} repeatedly collects fresh relative preferences from the evolving policy, allowing the steering direction to adapt as new behaviors emerge.

\subsection{Generative Policy Improvement}

Diffusion and flow-matching policies provide expressive multimodal action distributions~\citep{chi2023diffusion,lipman2023flow,black2024pi0}, motivating methods that steer generative policies using rewards or preferences. Reward-based approaches include DRaFT~\citep{clark2024draft}, DPPO~\citep{ren2025dppo}, and ORW-CFM-W2~\citep{fan2025onlineflow}, while preference-based approaches include Diffusion-DPO~\citep{wallace2024diffusion} and PC-Flow~\citep{wang2026pcflow}.

Most closely related is CFGRL~\citep{frans2025diffusion}, which interprets classifier-free guidance as a policy-improvement operator when the conditional-to-unconditional density ratio encodes an optimality factor. We build on this view but derive the signal from relative preferences and reconstruct the preference condition from newly generated trajectories at every iteration. This iterative structure is also related to self-improvement methods that repeatedly generate and retrain on selected samples~\citep{he2026sail}, but \methodabbr{} uses fresh human preferences to continually redirect the policy.

\section{Preliminaries}
\label{sec:preliminaries}

We consider a pretrained generative robot policy
$\pi_\theta(a_t \mid s_t,c)$, where $c$ is an optional discrete conditioning variable. Our formulation applies to diffusion~\citep{chi2023diffusion} and flow-matching policies~\citep{lipman2023flow,black2024pi0}. We denote the trajectory distribution induced by a policy as
\begin{equation}
    \tau \sim p_\pi(\tau \mid c).
\end{equation}
In \methodabbr{}, $c=1$ denotes the preference-conditioned branch and $c=\varnothing$ the unconditional branch.

Classifier-free guidance (CFG)~\citep{ho2022classifierfree} combines unconditional and conditional generative distributions. At the trajectory level, we use the idealized abstraction
\begin{equation}
\label{eq:cfg_prelim}
    p_w(\tau)
    \propto
    p(\tau)
    \left(
        \frac{p(\tau \mid c)}
             {p(\tau)}
    \right)^w,
\end{equation}
where $w \geq 0$ is the guidance strength. Increasing $w$ amplifies trajectories with a larger conditional-to-unconditional density ratio. CFGRL~\citep{frans2025diffusion} connects this ratio to policy improvement when it represents an optimality-related factor. We show how relative preference feedback induces such a factor in Section~\ref{sec:preference_as_evaluation}.

Eq.~\ref{eq:cfg_prelim} is an idealized trajectory-level abstraction used for our analysis; in practice, CFG is applied within the action-generative policy.

\section{Preference-Guided Policy Iteration}
\label{sec:method}

We formulate preference learning as \emph{conditional generative modeling}, where relative preference serves as a conditioning variable and CFG amplifies the induced conditional-to-unconditional density ratio.

\begin{figure*}[t]
    \centering
    \includegraphics[width=0.9\textwidth]{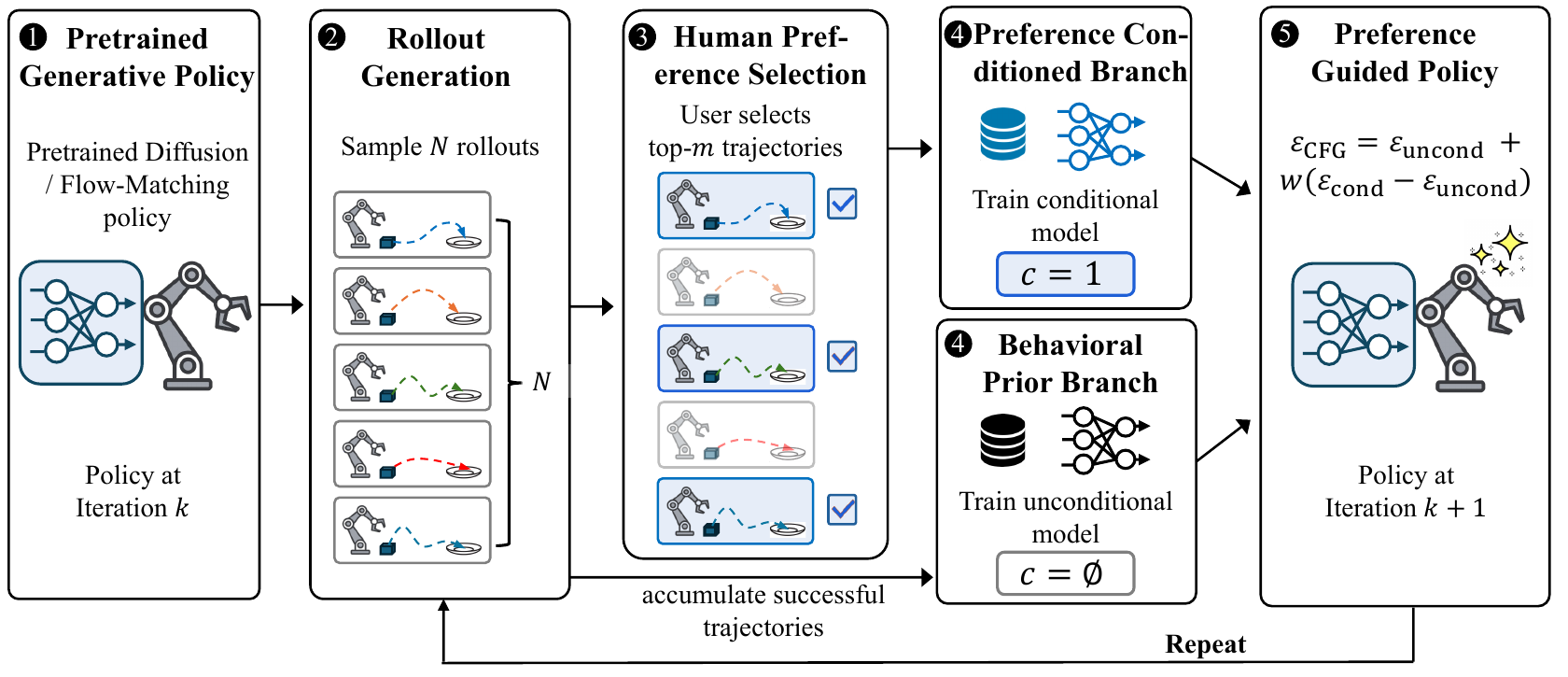}
    \caption{Overview of \methodabbr{}. Each iteration trains two branches, one conditioned on the top-$m$ rollouts and the other on all successful rollouts, and CFG combines the two branches to produce the next policy.}
    \label{fig:method_overview}
    \vspace{-1em} 
\end{figure*}

\subsection{Method Overview}
\label{sec:method_overview}

We illustrate the overall procedure of \methodabbr{} in Figure~\ref{fig:method_overview}. Starting from a pretrained generative policy, \methodabbr{} repeatedly alternates between relative preference evaluation and CFG-based policy improvement. At iteration $k$, let $p_k$ denote the trajectory distribution induced by the current policy.
We generate a batch of trajectories
\begin{equation}
    \mathcal{D}_k
    =
    \{\tau_1,\ldots,\tau_N\},
    \qquad
    \tau_i \sim p_k(\tau),
\end{equation}
from which the user selects the top-$m$ according to their preference.
We denote the selected subset by $\mathcal{D}_k^+$. We train the preference-conditioned branch $c=1$ on $\mathcal{D}_k^+$. For the unconditional branch, we maintain an accumulated dataset $\mathcal{D}_k^{\mathrm{acc}}$ of successful trajectories:
\begin{equation}
\mathcal{D}_k^{\mathrm{acc}}
=
\mathcal{D}_{k-1}^{\mathrm{acc}}
\cup
\operatorname{Success}(\mathcal{D}_k).
\end{equation}
The conditional branch captures the current preference direction, while the accumulated unconditional dataset preserves successful behavioral diversity for subsequent exploration. CFG combines the two branches to produce the policy for the next iteration, yielding the generate--select--guide--redeploy loop
$$
    p_k
    \rightarrow
    \mathcal{D}_k
    \rightarrow
    \mathcal{D}_k^+
    \rightarrow
    p_{k+1}.
$$

Conceptually, \methodabbr{} can be viewed as a trajectory-level form of \textbf{generalized policy iteration}~\citep{sutton2018reinforcement}: relative top-$m$ selection provides an implicit policy-evaluation signal, while CFG acts as a generative policy-improvement operator that amplifies preferred behavior. Because the generative policy can generalize beyond the finite preferred samples, previously rare behaviors may become exposed in the next rollout batch and serve as stepping stones for subsequent updates. Repeating this process can progressively shift the policy's effective support toward behaviors that were unlikely under the initial policy. We formalize the evaluation and improvement components in Sections~\ref{sec:preference_as_evaluation} and~\ref{sec:generative_improvement}, respectively.

\subsection{Relative Preference as an Optimality Signal}
\label{sec:preference_as_evaluation}

To interpret CFG as a policy-improvement operator, the conditional-to-unconditional density ratio must encode an optimality-related signal, as in CFGRL~\citep{frans2025diffusion}. In our setting, however, we do not assume access to numerical rewards or explicit optimality labels; the only supervision is the user's relative selection among trajectories sampled from the current policy. We therefore show that relative top-$m$ selection induces an order-preserving factor that can play the role of the optimality signal required by the CFG policy-improvement interpretation.

Let $p_k(\tau)$ be the current trajectory distribution and suppose user preference is induced by an unknown latent utility $U(\tau)$, with larger values corresponding to greater preference. The method never observes or estimates the numerical value of $U$.

We draw $N$ i.i.d.\ trajectories from $p_k$ and select the top $m$, with selection ratio $\rho=m/N$. Define
\begin{equation}
    g_k(u)
    \triangleq
    \Pr(c=1 \mid U(\tau)=u)
\end{equation}
as the probability that a trajectory of utility $u$ is selected.

Let
\begin{equation}
    F_k(u)
    =
    \Pr_{\tau\sim p_k}
    \left(U(\tau)\leq u\right)
\end{equation}
denote the utility CDF under the current policy. A trajectory with utility $u$ is selected among the top $m$ if at most $m-1$ of the other $N-1$ trajectories have greater utility. Therefore,
\begin{equation}
\label{eq:selection_probability}
    g_k(u)
    =
    \sum_{j=0}^{m-1}
    {N-1 \choose j}
    [1-F_k(u)]^j
    F_k(u)^{N-1-j}.
\end{equation}
Since $F_k(u)$ is non-decreasing in $u$, so is $g_k(u)$:
\begin{equation}
    u_1 \leq u_2
    \quad\Longrightarrow\quad
    g_k(u_1) \leq g_k(u_2).
\end{equation}
Thus, relative top-$m$ selection preserves the ordering induced by the latent utility.

To connect this selection signal to the preference-conditioned branch, let $p_k^+(\tau)$ denote the distribution of selected trajectories.
By Bayes' rule,
\begin{equation}
    p_k^+(\tau) =p_k(\tau \mid c=1)
    =
    \frac{
        p_k(\tau)\Pr(c=1\mid\tau)
    }{
        \Pr(c=1)
    }.
\end{equation}
Since
$\Pr(c=1\mid\tau)=g_k(U(\tau))$
and
$\Pr(c=1)=m/N=\rho$,
we obtain
\begin{equation}
\label{eq:preferred_distribution}
    p_k^+(\tau)
    =
    \frac{1}{\rho}
    p_k(\tau)
    g_k(U(\tau)).
\end{equation}
Thus, relative preference selection reweights the current trajectory distribution $p_k(\tau)$ by an order-preserving function $g_k(U(\tau))$ of the latent utility.

\subsection{CFG-Based Policy Improvement}
\label{sec:generative_improvement}

Having established the preference-induced signal $g_k(U(\tau))$, we now connect it to the CFG-based policy-improvement framework of CFGRL. Consider an idealized improvement step in which the unconditional branch represents $p_k(\tau)$ and the conditional branch represents $p_k^+(\tau)$. From Eq.~\ref{eq:preferred_distribution}, their density ratio is
\begin{equation}
    \frac{p_k^+(\tau)}{p_k(\tau)}
    =
    \frac{1}{\rho} g_k(U(\tau)).
\end{equation}

Let $\tilde{p}_{k,w}(\tau)$ denote the trajectory distribution obtained by applying
CFG with guidance strength $w$. By Eq.~\ref{eq:cfg_prelim},
\begin{align}
    \tilde{p}_{k,w}(\tau)
    &\propto
    p_k(\tau)
    \left(
        \frac{p_k^+(\tau)}
             {p_k(\tau)}
    \right)^w \\
    &\propto
    p_k(\tau)
    g_k(U(\tau))^w,
\end{align}
where the constant factor $\rho^{-w}$ is absorbed into the normalization.
Thus, CFG amplifies the preference-induced signal according to guidance strength $w$.

Because $g_k(U)$ is non-decreasing in $U$, increasing the CFG guidance strength cannot decrease the expected latent preference utility:
\begin{equation}
\label{eq:utility_improvement}
    \frac{\partial}{\partial w}
    \mathbb{E}_{\tau\sim\tilde{p}_{k,w}}
    \left[ U(\tau) \right]
    \geq 0.
\end{equation}
See Appendix~\ref{app:monotonic_improvement} for the derivation.
Since $\tilde{p}_{k,0}=p_k$, it follows that any $w\geq 0$ yields an expected latent utility no lower than that of the current policy.
In this sense, CFG acts as a policy-improvement operator using an optimality signal induced solely by relative trajectory preferences.

Our practical algorithm differs from this analysis in one respect: rather than using the current $p_k$ as the unconditional reference, we train the unconditional branch on the accumulated dataset $\mathcal{D}_k^{\mathrm{acc}}$ to preserve diversity. The monotonicity result therefore characterizes the idealized improvement direction induced by preference and CFG, while the effectiveness of the accumulated prior is evaluated empirically. Algorithm~\ref{alg:prefpi}
provides the complete preference-guided policy iteration procedure.

\section{Experimental Results} 

In this section, we aim to answer four questions: (1) Can \methodabbr{} steer pretrained large VLA policies beyond their initial effective support? (2) Can \methodabbr{} scale to real-world robot deployment? (3) Given the same preference-feedback budget, can \methodabbr{} achieve larger behavioral shifts than existing preference-learning and policy-steering baselines? (4) Does \methodabbr{} remain effective for conventional in-distribution preference refinement?

\subsection{Steering Large Vision-Language-Action Models} 
\label{sec:exp_pi_sim}

We evaluate \methodabbr{} using PI0.5-LIBERO~\citep{pmlr-v305-black25a}, a flow-matching VLA policy, on the LIBERO benchmark~\citep{liu2023libero}. We consider two preference objectives: \texttt{speed}, which favors faster task completion, and \texttt{detour}, which favors trajectories with larger vertical or horizontal displacement. We evaluate \texttt{speed} on three LIBERO-Goal tasks---\texttt{OpenDrawer}, \texttt{BowlToStove}, and \texttt{PushPlate}---and \texttt{detour} on \texttt{BowlToStove} and \texttt{PushPlate}. At each iteration, we collect 40 trajectories and select the top $40\%$ according to the corresponding preference criterion. Additional implementation details are provided in Appendix~\ref{app:vla_details}. 

\paragraph{PrefPI steers large VLA policies beyond their initial effective support.} 

\begin{figure*}[t] 
    \centering \includegraphics[width=1.0\textwidth]{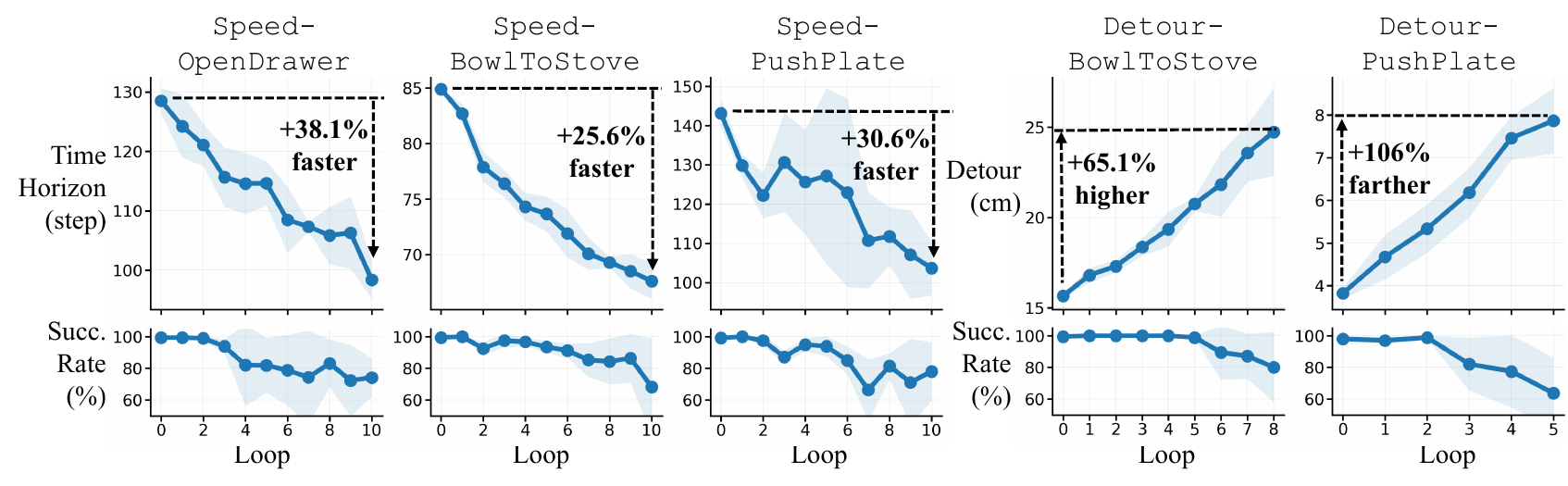} 
    \vspace{-1.3em}
    \caption{\methodabbr{} progressively steers the policy toward preferred behaviors. Results are averaged over three seeds. Top: preference metric; bottom: success rate.} 
    \label{fig:pi0.5_learning_curve} 
    \vspace{-0.5em}
\end{figure*} 

\begin{figure*}[t] 
    \centering \includegraphics[width=1.0\textwidth]{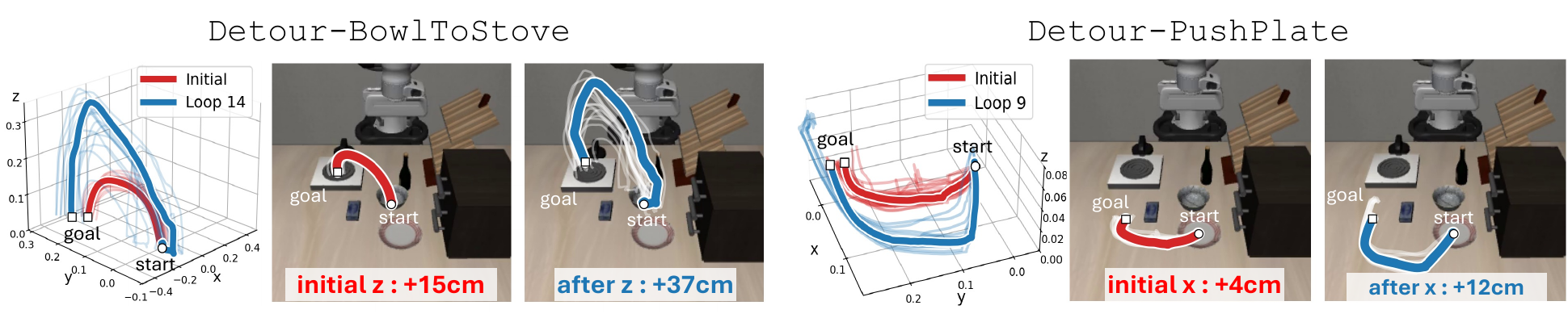} 
    \vspace{-0.5em}
    \caption{Trajectories before and after training are highly disjoint, indicating that the blue trajectories lie beyond the initial policy’s effective support.} \label{fig:detour_qualitative} 
\end{figure*} 

As shown in Figure~\ref{fig:pi0.5_learning_curve}, \methodabbr{} consistently steers the initial PI0.5 policy toward the specified preferences. On the \texttt{speed} tasks, execution speed improves by approximately $25\%$--$38\%$ within 10 iterations, corresponding to only 400 evaluated trajectories. On the \texttt{detour} tasks, the maximum bowl height increases by $65.1\%$ within 320 trajectories, while the plate is pushed $106\%$ farther along the preferred direction within 200 trajectories. Figure~\ref{fig:detour_qualitative} further shows that the adapted trajectories extend into regions largely disjoint from those visited by the initial policy. Notably, these behaviors emerge solely through iterative preference feedback, without demonstrations of the desired trajectories. In most experiments, success rates stay above $90\%$ over the early iterations, which already yield roughly half of the total shift. However, more aggressive steering comes with a $20\%$--$30\%$ decrease in task success rate, reflecting a trade-off between exploring behaviors beyond the initial policy distribution and preserving task reliability. We discuss this trade-off further in Section~\ref{sec:discussion}. 

\subsection{Real-World Preference-Guided Steering} 
\label{sec:exp_pi_real}

\begin{figure*}[t] 
    \centering \includegraphics[width=0.95\textwidth]{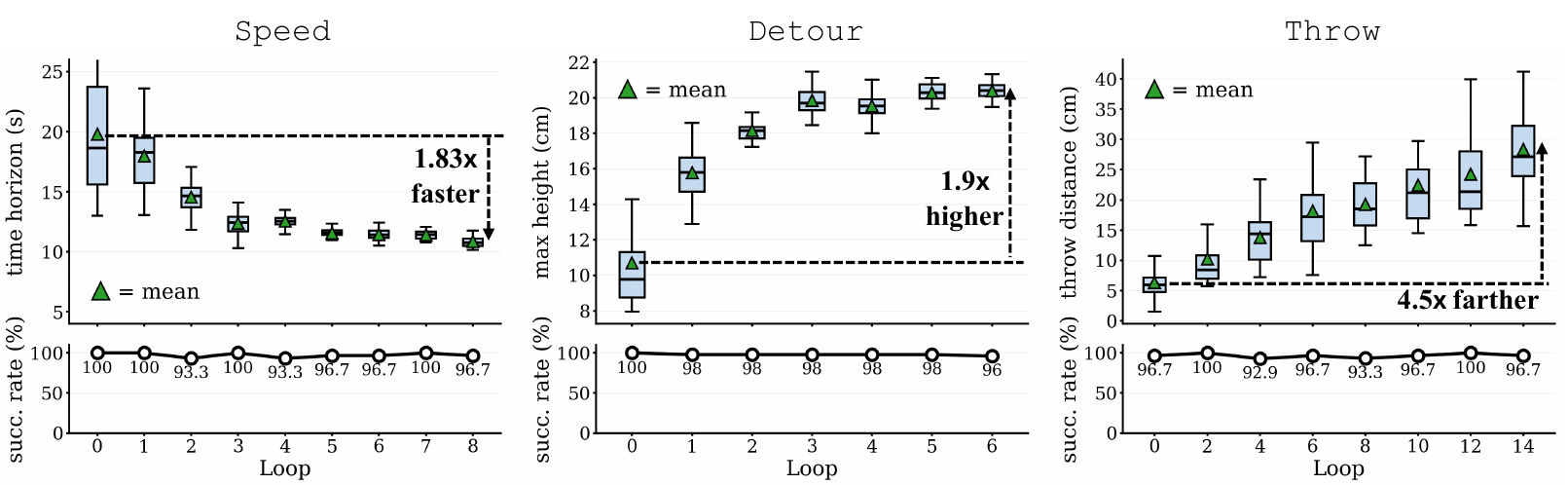} 
    \vspace{-0.5em}
    \caption{Real-world steering on three different preferences. \methodabbr{} improves the preference metric by $1.83\times$ (\texttt{speed}), $1.9\times$ (\texttt{detour}), and $4.5\times$ (\texttt{throw}), while success rates remain above $90\%$.} \label{fig:real_world_boxplot} 
    \vspace{-0.5em}
\end{figure*} 

We deploy \methodabbr{} on a real-world \texttt{CubePickAndPlace} task using a WidowX~\citep{trossenwidowx250} robot arm, where the robot must grasp a red cube and place it on a plate. Because zero-shot PI0.5 achieves a $0\%$ success rate, we first fine-tune it using 100 task demonstrations to obtain the initial policy. We then evaluate the same \texttt{speed} and \texttt{detour} preferences as in LIBERO, using 30 and 50 trajectories per iteration, respectively. We additionally consider a \texttt{throw} objective that favors longer throwing distances, using 30 trajectories per iteration. All other hyperparameters are kept identical to those used in the LIBERO experiments.

\paragraph{\methodabbr{} enables substantial real-world steering.} Interestingly, \methodabbr{} achieves larger behavioral changes with fewer interaction samples than in the simulated PI0.5 experiments while maintaining a high task success rate. The learning curves are shown in Figure~\ref{fig:real_world_boxplot}.

\begin{figure*}[t] 
    \centering       \includegraphics[width=1.0\textwidth]{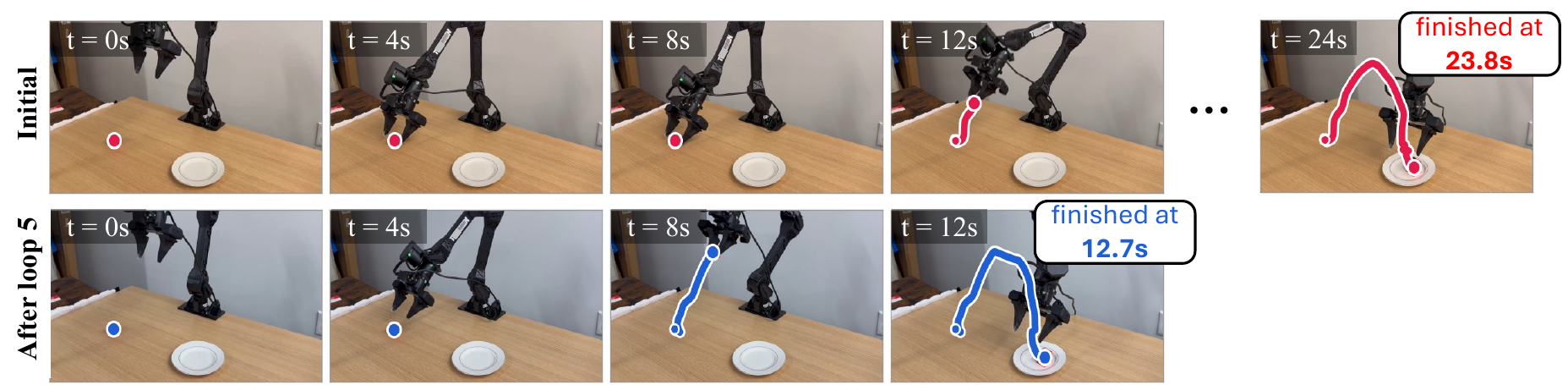} 
    \vspace{-0.5em}
    \caption{\methodabbr{} completes the task nearly \textbf{twice} as fast as the initial policy using only 150 preference-labeled trajectories (30 per iteration over 5 iterations).} 
    \label{fig:real_world_speed} 
\end{figure*}

Across all three objectives, \methodabbr{} produces substantial behavioral shifts: task execution becomes $1.83\times$ faster, maximum object height increases by $1.9\times$, and throwing distance increases by approximately $4.5\times$. Importantly, the \textbf{success rate remains above $90\%$} for most iterations across all three objectives, indicating that substantial steering can be achieved while largely preserving task competence. Representative executions for \texttt{detour}, \texttt{speed}, and \texttt{throw} are shown in Figures~\ref{fig:teaser}, \ref{fig:real_world_speed}, and~\ref{fig:throw_real}, respectively.

\subsection{OOD Steering under Matched Feedback Budgets}
\label{sec:baseline_ood}

We compare \methodabbr{} against representative preference-learning and policy-steering baselines on RoboMimic~\citep{robomimic2021} under matched preference-feedback budgets. We equalize the number of human preference labels across methods and evaluate all methods iteratively using fresh feedback from the current policy, allowing each method to exploit newly discovered behaviors as the policy distribution evolves. FDPP additionally collects 50 trajectories per iteration for its online RL update, resulting in approximately twice as many environment interactions as \methodabbr{}.

We compare against the following baselines:
\begin{itemize}
\item \textbf{Iterative Filtered Behavior Cloning (IFBC)} repeatedly fine-tunes the policy on preferred trajectories collected at each iteration, without classifier-free guidance.
\item \textbf{Direct Preference Optimization (DPO)}~\citep{rafailov2023direct, zhang2024grape} directly optimizes the policy from pairwise preference data.

\item \textbf{FDPP}~\citep{chen2025fdpp} follows an RLHF-style pipeline for diffusion policies, first learning a reward model from preference feedback and subsequently optimizing the policy using DPPO~\citep{ren2025dppo}.

\item \textbf{DSRL}~\citep{wagenmaker2025steering} performs reinforcement learning over the diffusion noise-sampling distribution to steer the behavior of a pretrained diffusion policy.

\end{itemize}

Additional implementation details, including our iterative implementations
and DPO stabilization, are provided in Appendix~\ref{app:baseline_details}.

\paragraph{\methodabbr{} achieves strong OOD steering under matched preference budgets.}

We consider three OOD steering tasks: \texttt{CanDropPos}, which favors larger directional displacement of the final can position; \texttt{CanSpeed}, which favors faster completion; and \texttt{LiftDetour}, which favors larger end-effector displacement during the approach. For each task, all methods start from the same diffusion policy pretrained on the corresponding RoboMimic human demonstration dataset.

\begin{figure*}[t]
\centering
\includegraphics[width=1.0\textwidth]{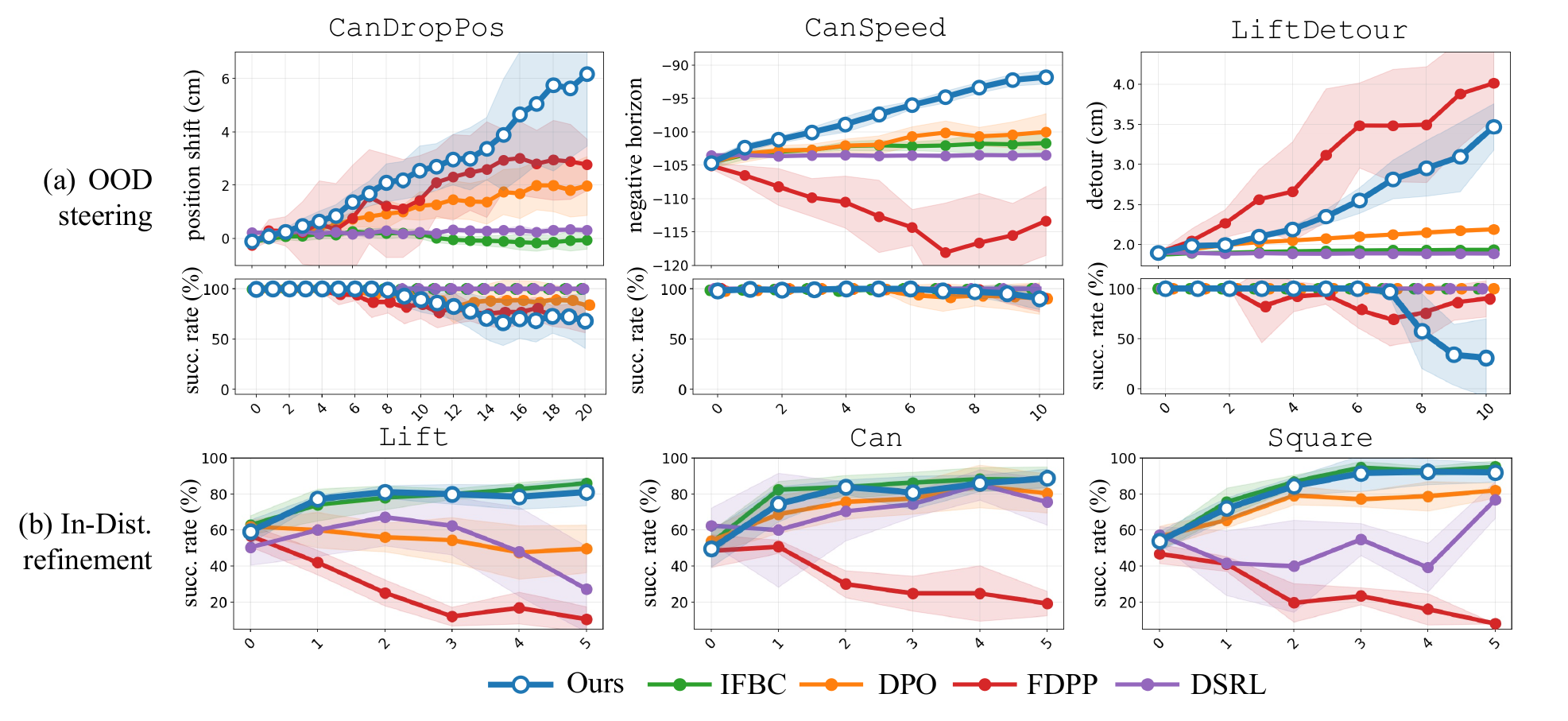}

\caption{Comparison with baselines, averaged over five seeds. \methodabbr{} achieves \textbf{(a)} the strongest OOD steering overall, \textbf{(b)} while matching the best baseline on in-distribution refinement.}
\vspace{-1em}
\label{fig:baseline_plot}
\end{figure*}

As shown in Figure~\ref{fig:baseline_plot}(a), \methodabbr{} achieves the strongest steering on \texttt{CanDropPos} and \texttt{CanSpeed}. On \texttt{LiftDetour}, FDPP achieves both a larger detour and a higher success rate than \methodabbr{}. However, FDPP is less consistent across tasks: it underperforms \methodabbr{} on the other two settings and fails to produce meaningful improvement on \texttt{CanSpeed}. At the same time, aggressive \texttt{LiftDetour} steering with \methodabbr{} reduces the success rate to roughly $30\%$, highlighting that large OOD behavioral shifts can come at a substantial cost in task reliability. We discuss this trade-off further in Section~\ref{sec:discussion}.

IFBC and DSRL exhibit little behavioral shift in these OOD steering tasks. IFBC directly trains on preferred trajectories sampled from the current policy and therefore primarily reinforces behaviors already observed in the collected data. In contrast, classifier-free guidance in \methodabbr{} can bias generation beyond the preferred samples themselves. Similarly, DSRL reweights the sampling process of the pretrained diffusion policy, which may limit how far it can move beyond behaviors represented by the initial policy. DPO produces a measurable steering effect, but the resulting shift is consistently smaller than that of \methodabbr{}.

\subsection{In-Distribution Preference Refinement}
\label{sec:baseline_in_dist}

We next consider the conventional preference-learning setting, where the desired behavior is already represented by the initial policy. Using the same baselines and matched preference-feedback protocol described above, we train diffusion policies on \texttt{Lift}, \texttt{Can}, and \texttt{Square}, intentionally stopping pretraining such that the initial policies achieve approximately $50\%$ task success. Preferences are determined solely by task success, with successful trajectories preferred over unsuccessful ones.

\paragraph{OOD steering does not come at the expense of in-distribution refinement.}
As shown in Figure~\ref{fig:baseline_plot}(b), \methodabbr{} improves task success across all three environments within five iterations. \methodabbr{} and IFBC perform best overall, with IFBC slightly outperforming \methodabbr{} on \texttt{Lift} and \texttt{Can}. This is consistent with Section~\ref{sec:generative_improvement}: when preferred behaviors are already in-distribution, $w=1$ recovers the preferred-trajectory distribution, so direct imitation can already be effective. CFG becomes more important for steering beyond the current policy distribution.

FDPP performs substantially worse, possibly because learning an accurate reward model from only 50 preference-labeled trajectories per iteration is difficult. Reward errors may then propagate through the online RL stage, reducing sample efficiency.

\section{Discussion: The Role of Prior Diversity}
\label{sec:discussion}

\begin{wrapfigure}{r}{0.3\textwidth}
    \centering
    \vspace{-2em}
    \includegraphics[width=\linewidth]{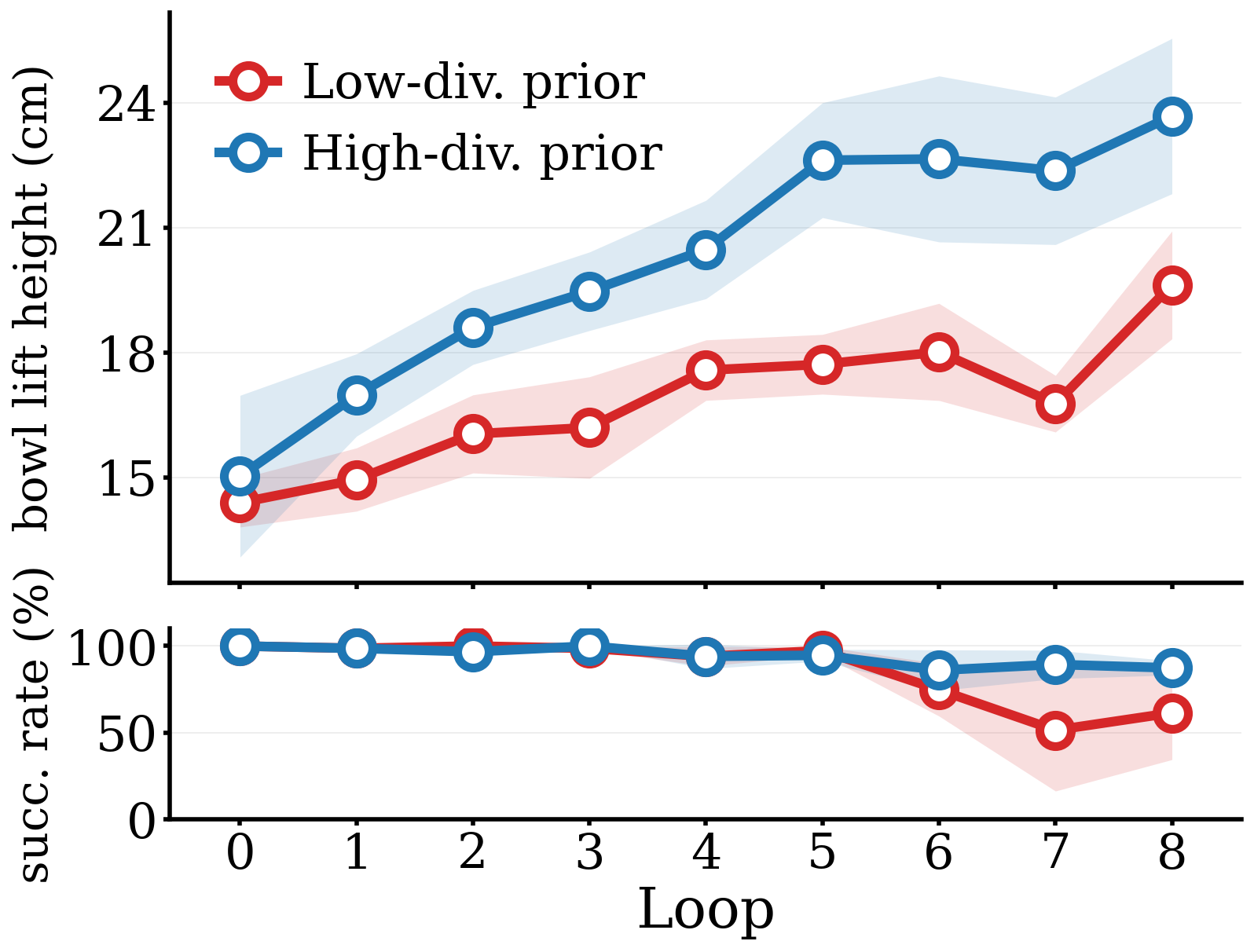}
    \caption{\methodabbr{} benefits from a high-diversity prior.}
    \label{fig:diversity_control_exp}
    \vspace{-1em}
\end{wrapfigure}

Across our experimental results, we observe a consistent trend: greater behavioral diversity in the initial policy leads to stronger steering with \methodabbr{}. For instance, \texttt{LiftDetour} in Section~\ref{sec:baseline_ood}, where \methodabbr{} showed the largest drop in task success, had very limited behavioral diversity in the initial policy. Across 250 rollouts, the mean detour was 1.90~cm with a standard deviation of only 0.022~cm, corresponding to a relative std.\ of $1.1\%$. In contrast, the initial policies for the \texttt{detour} experiments in Sections~\ref{sec:exp_pi_sim} and~\ref{sec:exp_pi_real} exhibited substantially greater behavioral diversity, with relative std.\ values of $6.8\%$ and $27.2\%$, respectively. In both cases, \methodabbr{} achieved strong steering while maintaining high task success.

These observations suggest that prior diversity may benefit \methodabbr{}. We argue that this benefit arises through two complementary mechanisms.
First, diversity may improve generalization during CFG-driven exploration.
Because CFG shifts the policy toward increasingly underrepresented behaviors, successful steering requires the policy to retain task competence in these newly explored regions.
A broader prior may therefore allow larger shifts through better generalization to newly explored behaviors.
Second, diversity may strengthen the directional signal provided by CFG.
CFG amplifies relative differences between the preference-conditioned and unconditional distributions; when the prior contains behaviorally distinct solutions, these relative differences may provide a clearer steering direction than when all sampled behaviors are narrowly clustered.

\paragraph{Greater prior diversity enables stronger steering.}
To confirm this, we evaluate \methodabbr{} using two different PI0.5-based priors on the \texttt{Detour-BowlToStove} task over three seeds. The two priors are trained from 100 synthetically generated demonstrations with different degrees of diversity. Both demonstration sets are centered at a transport height of 15~cm, but are evenly distributed over different ranges, 12.5--17.5~cm and 14--16~cm for the high- and low-diversity priors, respectively, while all other factors are held fixed.

As shown in Figure~\ref{fig:diversity_control_exp}, the high-diversity prior reaches approximately 24~cm while maintaining above $85\%$ success, whereas the low-diversity prior reaches only approximately 20~cm and drops to roughly $50\%$ success. Our controlled experiment confirms that greater initial behavioral diversity enables both larger preference-driven steering and better task reliability.

\section{Conclusion}
\label{sec:conclusion}

We introduced \methodabbr{}, a lightweight preference-guided framework for steering generative visuomotor policies beyond their initial effective support. Across simulation and real-world experiments, \methodabbr{} achieves substantial behavioral shifts using only relative preferences. As discussed in Section~\ref{sec:discussion}, a key limitation is that steering performance depends on the diversity of the initial policy, motivating future work on improving robustness under narrow behavioral priors.

\bibliography{iclr2027_conference}
\bibliographystyle{iclr2027_conference}

\newpage

\appendix

\section{Algorithm}
\label{sec:algorithm}

\begin{algorithm}[h!]
\caption{\methodabbr{}: Preference-Guided Policy Iteration}
\label{alg:prefpi}

\begin{algorithmic}[1]
\Require Pretrained generative policy $\pi_0$, number of iterations $K$,
rollout batch size $N$, number of preferred trajectories $m$,
guidance strength $w$

\State Initialize accumulated successful dataset
$\mathcal{D}_{-1}^{\mathrm{acc}} \gets \emptyset$

\For{$k = 0,1,\ldots,K-1$}

    \State Let $p_k(\tau) \equiv p_{\pi_k}(\tau)$ denote the trajectory distribution induced by $\pi_k$

    \State Generate $N$ trajectories:
    \[
        \mathcal{D}_k
        =
        \{\tau_i\}_{i=1}^{N},
        \qquad
        \tau_i \sim p_k(\tau)
    \]

    \State Obtain the user's relative preference over $\mathcal{D}_k$

    \State Select the top-$m$ preferred trajectories:
    \[
        \mathcal{D}_k^{+}
        \gets
        \operatorname{TopM}(\mathcal{D}_k)
    \]

    \State Update the accumulated successful dataset:
    \[
        \mathcal{D}_k^{\mathrm{acc}}
        \gets
        \mathcal{D}_{k-1}^{\mathrm{acc}}
        \cup
        \operatorname{Success}(\mathcal{D}_k)
    \]

    \State Train the unconditional and preference-conditioned branches:
    \[
        (\pi_{k,\varnothing}, \pi_{k,1})
        \gets
        \operatorname{Train}
        \left(
            \pi_k;
            \mathcal{D}_k^{\mathrm{acc}},
            \mathcal{D}_k^{+}
        \right)
    \]

    \State Construct the next policy using classifier-free guidance:
    \[
        \pi_{k+1}
        \gets
        \operatorname{CFG}
        \left(
            \pi_{k,\varnothing},
            \pi_{k,1};
            w
        \right)
    \]

\EndFor

\State \Return $\pi_K$

\end{algorithmic}
\end{algorithm}

\begin{figure*}[t] 
    \centering \includegraphics[width=0.95\textwidth]{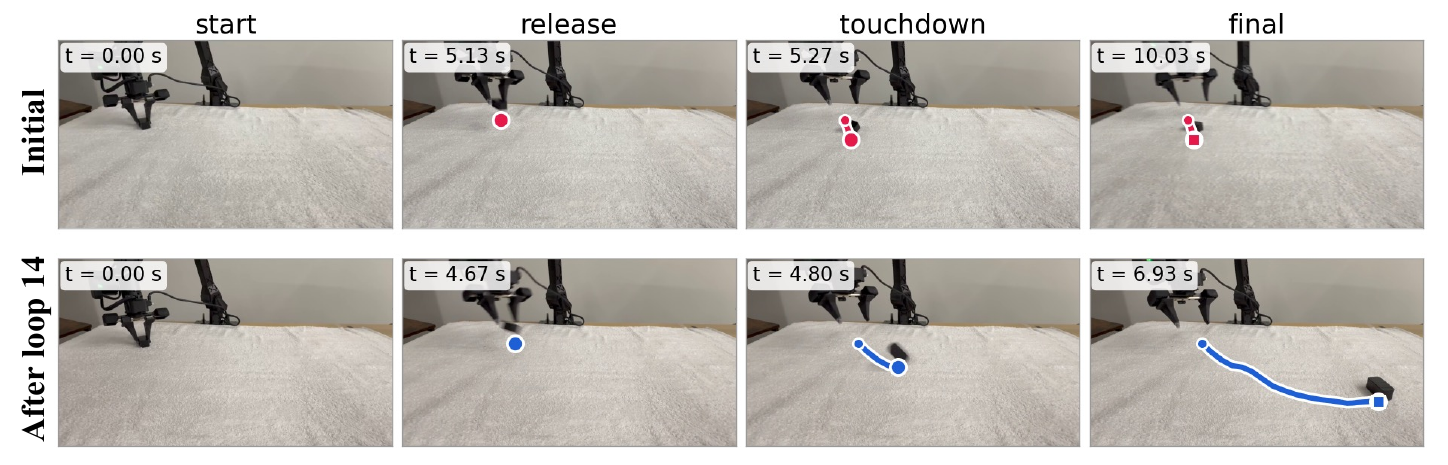} 
    \caption{Example \texttt{throw} rollouts. The initial policy barely throws the object, while by iteration 14, \methodabbr{} achieves a much longer throwing distance. A towel absorbs impact in the landing area.} 
    \label{fig:throw_real}
\end{figure*}

\section{Monotonic Improvement with Guidance Strength}
\label{app:monotonic_improvement}

Here we provide the derivation of the monotonic improvement result stated in
Section~\ref{sec:generative_improvement}. Recall that CFG induces the trajectory
distribution
\begin{equation}
\tilde{p}_{k,w}(\tau)
\propto
p_k(\tau) g_k(U(\tau))^w.
\end{equation}
Writing this distribution explicitly,
\begin{equation}
\tilde{p}_{k,w}(\tau)
=
\frac{
p_k(\tau) g_k(U(\tau))^w
}{
Z_k(w)
},
\end{equation}
where
\begin{equation}
Z_k(w)
=
\mathbb{E}_{\tau\sim p_k}
\left[
g_k(U(\tau))^w
\right].
\end{equation}

Define the expected latent preference utility under the guided distribution as
\begin{equation}
J_k(w)
=
\mathbb{E}_{\tau\sim\tilde{p}_{k,w}}
\left[
U(\tau)
\right].
\end{equation}
Under standard integrability conditions, we show that
\begin{equation}
\frac{\partial J_k(w)}{\partial w}
\geq 0.
\end{equation}

Taking the logarithm of the guided distribution gives
\begin{equation}
\log \tilde{p}_{k,w}(\tau)
=
\log p_k(\tau)
+
w\log g_k(U(\tau))
-
\log Z_k(w).
\end{equation}
Differentiating with respect to $w$,
\begin{equation}
\frac{\partial}{\partial w}
\log \tilde{p}_{k,w}(\tau)
=
\log g_k(U(\tau))
-
\frac{\partial}{\partial w}\log Z_k(w).
\end{equation}
The normalization term satisfies
\begin{align}
\frac{\partial}{\partial w}\log Z_k(w)
&=
\frac{1}{Z_k(w)}
\frac{\partial Z_k(w)}{\partial w} \\
&=
\frac{
\mathbb{E}_{\tau\sim p_k}
\left[
g_k(U(\tau))^w
\log g_k(U(\tau))
\right]
}{
Z_k(w)
} \\
&=
\mathbb{E}_{\tau\sim\tilde{p}_{k,w}}
\left[
\log g_k(U(\tau))
\right].
\end{align}
Therefore,
\begin{equation}
\frac{\partial}{\partial w}
\log \tilde{p}_{k,w}(\tau)
=
\log g_k(U(\tau))
-
\mathbb{E}_{\tau\sim\tilde{p}_{k,w}}
\left[
\log g_k(U(\tau))
\right].
\end{equation}

Using the score-function identity,
\begin{align}
\frac{\partial J_k(w)}{\partial w}
&=
\mathbb{E}_{\tau\sim\tilde{p}_{k,w}}
\left[
U(\tau)
\frac{\partial}{\partial w}
\log \tilde{p}_{k,w}(\tau)
\right] \\
&=
\operatorname{Cov}_{\tau\sim\tilde{p}_{k,w}}
\left[
U(\tau),
\log g_k(U(\tau))
\right].
\end{align}

Since $g_k(U)$ is non-decreasing in $U$, $\log g_k(U)$ is also
non-decreasing wherever $g_k(U)>0$. Two non-decreasing functions of the same
random variable have non-negative covariance. Therefore,
\begin{equation}
\frac{\partial J_k(w)}{\partial w}
=
\operatorname{Cov}_{\tau\sim\tilde{p}_{k,w}}
\left[
U(\tau),
\log g_k(U(\tau))
\right]
\geq 0.
\end{equation}

Hence, under the idealized distributional assumptions of
Section~\ref{sec:generative_improvement}, increasing the CFG guidance strength
cannot decrease the expected latent preference utility. Since
$\tilde{p}_{k,0}=p_k$, this also implies
\begin{equation}
\mathbb{E}_{\tau\sim\tilde{p}_{k,w}}
[U(\tau)]
\geq
\mathbb{E}_{\tau\sim p_k}
[U(\tau)],
\qquad w\geq 0.
\end{equation}

\section{Additional Details for VLA Steering}
\label{app:vla_details}

\paragraph{Model and fine-tuning.}
We initialize all experiments from the LIBERO-finetuned PI0.5 checkpoint (\texttt{pi05\_libero}). The policy consists of a PaliGemma backbone with a SigLIP So400m/14 vision encoder and Gemma-2B language model, together with a Gemma-300M action expert. We apply LoRA~\citep{hu2022lora} to the Gemma-2B backbone and action expert, while fully fine-tuning the SigLIP vision encoder and the action/time projection layers. Thus, our adaptation should be viewed as LoRA on the language and action backbones combined with full fine-tuning of the vision and projection modules. Overall, approximately $467$M parameters, or $13.7\%$ of the model, are trainable.

\begin{table}[h!]
\centering
\small
\caption{Model and fine-tuning configuration for PI0.5.}
\label{tab:vla_model}
\begin{tabular}{ll}
\toprule
\textbf{Hyperparameter} & \textbf{Value} \\
\midrule
Base policy & PI0.5-LIBERO (\texttt{pi05\_libero}) \\
VLM backbone & SigLIP So400m/14 + Gemma 2B \\
Action expert & Gemma 300M \\
Action dim. / horizon & 32 (7 used) / 10 \\
Max. prompt length & 200 tokens \\
Compute dtype & bfloat16 \\
LoRA rank / $\alpha$ (Gemma 2B) & 16 / 16 \\
LoRA rank / $\alpha$ (action expert) & 32 / 32 \\
Trainable parameters & 467M (13.7\%) \\
\bottomrule
\end{tabular}
\end{table}

\paragraph{CFG conditioning and data construction.}
We implement the conditioning variable through the natural-language task prompt. For the unconditional branch, $c=\varnothing$, we use the original task instruction. For the preference-conditioned branch, $c=1$, we prepend a special token, \texttt{[cfg]}, to the same instruction. This allows both branches to share the same VLA architecture without introducing an additional conditioning input.

At each iteration, we collect 40 rollouts and rank successful trajectories according to the corresponding preference objective. The top $40\%$ are added to the preferred set. Preferred trajectories are maintained in a FIFO buffer of size 40, while successful unconditional trajectories are accumulated across iterations. During training, unconditional and preference-conditioned samples are drawn with a unconditional-to-conditional sampling ratio of $0.4{:}1$. Sampling weights are assigned per episode, with frames sampled uniformly within each episode.

\begin{table}[t]
\centering
\small
\caption{Optimization and online adaptation settings.}
\label{tab:vla_training}
\begin{tabular}{ll}
\toprule
\textbf{Hyperparameter} & \textbf{Value} \\
\midrule
Optimizer & AdamW \\
Adam $\beta_1,\beta_2$ & $0.9, 0.95$ \\
Peak learning rate & $5\times10^{-5}$ \\
LR schedule & 20-step warmup, then constant \\
Weight decay & $10^{-10}$ \\
Gradient clipping & Global norm 1.0 \\
Batch size & 12 \\
Training steps / iteration & 3,000 \\
Rollouts / iteration & 40 \\
Preferred fraction & Top $40\%$ of successful rollouts \\
Preferred buffer & FIFO, 40 episodes \\
Unconditional : conditional sampling & $0.4{:}1$ \\
Checkpoint initialization & Previous iteration \\
Hardware & 1$\times$ NVIDIA L40S \\
\bottomrule
\end{tabular}
\end{table}

\paragraph{Evaluation and inference.}
We use 10 flow-matching denoising steps and apply classifier-free guidance at every sampling step with guidance scale $w=5.0$:
\begin{equation}
    v_{\mathrm{CFG}}
    =
    v_{\mathrm{uncond}}
    +
    w\left(
    v_{\mathrm{cond}}
    -
    v_{\mathrm{uncond}}
    \right).
\end{equation}
Each policy is evaluated for 40 episodes per iteration. The policy observes agent-view and wrist-camera images rendered at $256\times256$ resolution and resized with padding to $224\times224$. We replan every 5 environment steps using a 10-step action chunk, and episodes are limited to 300 control steps following 10 warm-up steps.

\paragraph{Detour scoring.}
For the vertical-detour experiment, we rank successful rollouts by the maximum bowl height attained while the object is grasped. Thus, the preference signal directly favors trajectories that transport the object along increasingly elevated paths rather than merely improving success on the original task. Analogous geometric scores are used for the horizontal-detour experiments.

\clearpage
\section{Baseline Implementation Details}
\label{app:baseline_details}

We make substantial efforts to obtain strong implementations of all baselines.
In particular, we adapt IFBC, DPO, DSRL, and FDPP to the same iterative
preference-feedback setting used by \methodabbr{}. This is important for the
OOD steering experiments in Section~\ref{sec:baseline_ood}, where the policy
distribution can change substantially over the course of adaptation.
Unless otherwise noted, all methods receive the same number of
preference-labeled trajectories at each iteration.

\paragraph{Iterative adaptation protocol.}
IFBC naturally admits an iterative implementation: at each iteration, the
current policy generates a new batch of trajectories, the preferred subset is
selected, and the policy is fine-tuned on these trajectories.
For DPO, we similarly collect fresh preferred--unpreferred pairs from the
current policy at every iteration and continue optimizing the DPO objective,
warm-starting from the policy obtained in the previous iteration.

We also make DSRL and FDPP iterative, including their reward-learning stage.
Both methods rely on a learned reward function to guide policy optimization.
If this reward model were trained only on trajectories from the initial policy,
its predictions could become unreliable as policy optimization discovers
behaviors far from its original training distribution. This issue is
particularly relevant in our OOD steering setting: a policy may discover a
more preferred behavior through exploration, while a reward model trained only
on the initial distribution may not reliably evaluate it. We therefore update
the reward model at every iteration using newly collected preference feedback,
allowing its training distribution to evolve together with the policy. This
gives DSRL and FDPP the same opportunity as \methodabbr{} to exploit newly
discovered behaviors.

\paragraph{Interaction budgets for DSRL and FDPP.}
DSRL can perform its policy update using an offline dataset, so we reuse the
same preference-labeled trajectories used to train its reward model for the
subsequent RL update. Thus, DSRL uses no additional environment interaction
beyond the trajectories counted toward the preference-feedback budget.

FDPP, in contrast, performs policy optimization with DPPO~\citep{ren2025dppo}, an online
PPO-based algorithm. After fitting the reward model from 50 preference-labeled
trajectories at each iteration, we therefore collect an additional 50
trajectories from the current policy for its online RL update. Consequently,
FDPP receives the same number of preference labels as all other methods but
uses approximately twice as many environment interactions. We allow these
additional rollouts so that FDPP is not disadvantaged by restricting the
online interaction required by its original optimization procedure.
\subsection{Diffusion-DPO Baseline and Stabilization}
\label{sec:dpo}

We implement DPO as an iterative Diffusion-DPO
baseline~\citep{wallace2024diffusion}. At each iteration, the current policy
generates a new batch of trajectories, from which preferred and unpreferred
samples are selected using the same ranking criterion as \methodabbr{}. The
policy is then fine-tuned and used to collect the next batch. We use the same
base checkpoint and optimization settings as \methodabbr{} wherever
applicable.

\paragraph{Diffusion-DPO objective.}
Let $m_\theta^{w}$ and $m_\theta^{l}$ denote the denoising MSEs of the
preferred and unpreferred samples under the current policy, and
$m_{\mathrm{ref}}^{w}$ and $m_{\mathrm{ref}}^{l}$ those under a frozen
reference policy. Diffusion-DPO minimizes
\begin{equation}
    \mathcal{L}_{\mathrm{DPO}}
    =
    -\mathbb{E}
    \left[
        \log \sigma
        \left(
            \beta(\Delta^{w}+\Delta^{l})
        \right)
    \right],
    \qquad
    \Delta^{w}=m_{\mathrm{ref}}^{w}-m_\theta^{w},
    \quad
    \Delta^{l}=m_\theta^{l}-m_{\mathrm{ref}}^{l}.
    \label{eq:dpo_vanilla}
\end{equation}

Directly applying this objective in our iterative setting was unstable.
In particular, the objective can be improved by increasing the denoising error
on unpreferred samples rather than improving preferred ones, a failure mode
related to excessive suppression of rejected samples observed in prior
work~\citep{razin2025unintentional,cho2025rethinking}. We therefore use the
following stabilization measures.

\paragraph{Capped unpreferred term.}
We cap the contribution of the unpreferred term and penalize increases beyond
the cap:
\begin{equation}
    \mathcal{L}
    =
    -\mathbb{E}
    \left[
        \log \sigma
        \left(
            \beta
            \left[
                \Delta^{w}
                +
                \min(\Delta^{l},\kappa)
            \right]
        \right)
    \right]
    +
    \lambda
    \mathbb{E}
    \left[
        \mathrm{SmoothL1}
        \left(
            \max(\Delta^{l}-\kappa,0)
        \right)
    \right],
    \label{eq:dpo_capped}
\end{equation}
where $\kappa=0.02$ and $\lambda=1$. This prevents optimization from
indefinitely increasing the error of unpreferred samples.

\paragraph{Iterative reference and sample buffers.}
At iteration $k$, the previous policy $\theta_{k-1}$ is used both to
initialize the trainable policy and as the frozen reference policy.
We maintain a FIFO preferred buffer of 40 episodes containing the top $40\%$
of successful trajectories and a FIFO negative buffer of 50 episodes populated
with the lowest-ranked non-preferred trajectories. Preferred and unpreferred
samples are randomly paired during training.

\paragraph{Optimization.}
We use $\beta=1$, gradient-norm clipping at $1.0$, and exponential moving
average (EMA) weights. These modifications substantially improved the
stability of the iterative DPO baseline in our experiments.

\end{document}